\documentclass[conference]{IEEEtran}
\usepackage{cite}
\usepackage{amsmath,amssymb,amsfonts}
\usepackage{algorithmic}
\usepackage{graphicx}
\usepackage{textcomp}
\usepackage{xcolor}
\usepackage[utf8]{inputenc}
\usepackage[T1]{fontenc} 
\usepackage[polish]{babel} %
\def\BibTeX{{\rm B\kern-.05em{\sc i\kern-.025em b}\kern-.08em
    T\kern-.1667em\lower.7ex\hbox{E}\kern-.125emX}}

\usepackage{multirow}

\begin{document}

\title{Improving Object Detection Quality in Football Through Super-Resolution Techniques}
% {\footnotesize \textsuperscript{*}Note: Sub-titles are not captured in Xplore and
% should not be used}
% \thanks{Identify applicable funding agency here. If none, delete this.}

% Enhancing Object Detection in Football Through Advanced super resolution Methods: An Empirical Study
% Breaking the Limits: super resolution's Role in Improving Object Detection Accuracy in Football

%\author{\IEEEauthorblockN{Anonymous Submission}}

\author{
\IEEEauthorblockN{
    Karolina Seweryn\IEEEauthorrefmark{1}, Gabriel Chęć\IEEEauthorrefmark{2}, Szymon Łukasik\IEEEauthorrefmark{2}\IEEEauthorrefmark{1}, Anna Wróblewska\IEEEauthorrefmark{3}
}
\IEEEauthorblockA{
    \IEEEauthorrefmark{1}NASK - National Research Institute\\
    \IEEEauthorrefmark{2}AGH University of Science and Technology\\
    \IEEEauthorrefmark{3}Warsaw University of Technology
}
}
\maketitle

%\author{\IEEEauthorblockN{Karolina Seweryn}
%\IEEEauthorblockA{\textit{NASK - National Research Institute} \\
%\textit{name of organization (of Aff.)}\\
%karolina.seweryn@nask.pl}
%\and
%\IEEEauthorblockN{Gabriel Chęć}
%\IEEEauthorblockA{\textit{AGH University of Science and Technology}
%}
%\and
%\IEEEauthorblockN{Szymon Łukasik}
%\IEEEauthorblockA{\textit{AGH University of Science and Technology} \\
%\textit{NASK - National Research Institute}
%}
%\and
%\IEEEauthorblockN{Anna Wróblewska}
%\IEEEauthorblockA{\textit{Warsaw University of Technology } \\
%}
%}

\begin{abstract}
This study explores the potential of super-resolution techniques in enhancing object detection accuracy in football. Given the sport's fast-paced nature and the critical importance of precise object (e.g. ball, player) tracking for both analysis and broadcasting, super-resolution could offer significant improvements. We investigate how advanced image processing through super-resolution impacts the accuracy and reliability of object detection algorithms in processing football match footage.

Our methodology involved applying state-of-the-art super-resolution techniques to a diverse set of football match videos from SoccerNet, followed by object detection using Faster R-CNN. The performance of these algorithms, both with and without super-resolution enhancement, was rigorously evaluated in terms of detection accuracy. % under varying conditions (e.g., lighting, weather).

The results indicate a marked improvement in object detection accuracy when super-resolution preprocessing is applied. The improvement of object detection through the integration of super-resolution techniques yields significant benefits, especially for low-resolution scenarios, with a notable 12\% increase in mean Average Precision (mAP) at an IoU (Intersection over Union) range of 0.50:0.95 for 320x240 size images when increasing the resolution fourfold using RLFN. As the dimensions increase, the magnitude of improvement becomes more subdued; however, a discernible improvement in the quality of detection is consistently evident.
%For images of size 320x240, adding 4 times RLFN resulted in mAP increase of 10\%. 
Additionally, we discuss the implications of these findings for real-time sports analytics, player tracking, and the overall viewing experience. The study contributes to the growing field of sports technology by demonstrating the practical benefits and limitations of integrating super-resolution techniques in football analytics and broadcasting.

\end{abstract}


\section{Introduction}

The tracking of players and the ball in football matches is a critical aspect of sports analytics, significantly impacting the tactical elements of the game. Player tracking data provides valuable insights into individual and team performance, enabling coaches to analyze movements, formations, and strategies more effectively~\cite{{survey-on-tracking}, survey-tracking-soccer}. For instance, tracking data can be used to assess player fitness, monitor fatigue, and reduce the risk of injury by understanding players' physical demands during a match. Furthermore, ball-tracking technology plays a key role in enhancing the spectator experience, contributing to more accurate and fair decisions through technologies like goal-line technology and VAR (Video Assistant Referee). 

Although new articles on tracking football players and the ball are published annually~\cite{survey-tracking-soccer, multi-camera-tracking-komorowski, deepplayer, self-supervised-small, SoccerNet-Tracking}, the precise tracking of small objects like a football ball remains a challenge. These advanced tracking solutions frequently depend on the availability of high-definition video input to ensure appropriate results. However, a significant portion of football clubs, ranging from grassroots clubs to professional organizations, struggle with the provision of such high-quality data. This situation is not solely a consequence of the prohibitive costs associated with state-of-the-art recording equipment but is also exacerbated by the prevalent use of drones and portable devices that, while offering versatility and convenience, often compromise on video quality. Additionally, the widespread practice of compressing video content for distribution on platforms like YouTube or other streaming services further degrades the fidelity of footage available for analytical purposes, presenting substantial hurdles to the deployment and effectiveness of sophisticated tracking models in real-world settings.
%Unfortunately, many football clubs or organizations cannot provide such high-quality data, which complicates the implementation of these models. It is partly due to device costs but also widespread use of drone/portable devices for recording video materials and compression of videos for their publication e.g. on YouTube or other video streaming services.

In this paper, we consider the impact of super-resolution methods to improve the quality of object detection models while reducing the requirements for input data quality in context of football analytics. Modern object detection models combined with super-resolution techniques that perform well with lower-resolution videos would reduce costs for teams, as they would not need to buy expensive equipment, decrease the amount of data that needs to be stored, and make it easier to use these technologies.

\section{Related Work}

% Object Detection/Tracking in Sports Analytics | Image Quality Enhancement for Sports Broadcasting |Challenges in Object Detection under Low Resolution or Challenging Conditions:

Initial approaches to extracting tracking data relied on views from multiple cameras~\cite{multi-camera,multi-camera2, multi-camera-sport, multi-camera-tracking-komorowski}. Thanks to this, machine learning algorithms were able to accurately locate e.g. football players. Unfortunately, installing many cameras is associated with high purchase costs and problems in maintaining infrastructure, including many technological challenges, such as multi-camera calibration and object re-identification~\cite{multi-camera2}. More and more latest models find the coordinates of players and the ball based on just one camera~\cite{single-camera,single-multi-camera}. Still, locating small objects like the ball remains a challenge. The ball might not be visible due to a wide camera view, occlusion, false detection, lighting variation, and different video quality with varying frame rates (various ball speeds)~\cite{soccer-tracking-survey}. Also, interactions between players, and players and the ball can cause complex problems. 
% ktos zauwazyl, ze... i postanowuil zastosowac rozwiazanie X
Additionally, solutions for detecting players and a ball often rely on high-resolution input videos or zoom-in views and their trajectory imputations, i.e. inferring their locations based on adjacent frames in which their detections were feasible~\cite{ball-trajectory}.
Tracking in a football match can be classified as a multi-object/target tracking task, and it remains challenging due to factors like abrupt object appearance differences and even severe object shadings and obscurations~\cite{multi-object-tracking}.

\begin{figure*}[!htbp]
\centerline{\includegraphics[width=\textwidth]{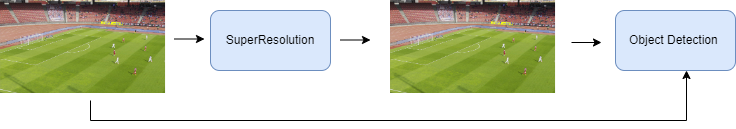}}
\caption{Experiment procedure.}
\label{fig:experiment-diagram}
\end{figure*}

% rozwiniecie jak to robia, na czym bazuja najnowsze rozwiazania
The latest solutions work with neural architectures to improve their ability to capture and detect small objects, e.g. adding focal loss to YOLOv7~\cite{tiny-ball}, or utilizing neural radiance fields~\cite{tiny-nerf}.

% podobny problem zostal poruszony w niektorych pracach i co zrobili
A similar issue of tracking football players in low-quality videos has been addressed in the works ~\cite{self-supervised-small} and ~\cite{low-quality-tracking}. In both cases, the researchers focus on adapting systems to challenging visual conditions typical of football broadcasts. The authors of~\cite{self-supervised-small} adapt advanced multi-object tracking systems for use with low-quality videos. At the same time, the other research concentrates on detecting and tracking small, less visible players without the need for manual data annotation. All those approaches highlight the increasing capability of sports analytics to handle visual challenges in football.

% super-resolution Techniques in Image Processing | Review Articles on the Current State of super-resolution and Object Detection:
In recent years, super-resolution approaches have been significantly improved~\cite{super-res-survey1}. In~\cite{super-res-survey2}, the authors introduced a taxonomy of super-resolution approaches, including reconstruction-based, interpolation-based, learning-based, and transformer-based algorithms. 
Currently, a bunch of research cope with applying and adapting super-resolution techniques to many detection tasks, such as satellite imagery~\cite{satellite-res}, underwater object detection~\cite{underwater-res}, and occluded small commodities~\cite{commodities-res}. 
As far as sports analytics is concerned, we have found only one research testing super-resolution techniques in preparing a dataset for ball detection and tracking in tennis matches~\cite{tennis-res}. 

% dodac o wplywie SR na detekcje 
% na innych zbiorach czy sprawdzali
% analiza super-resolution w kontekscie innego zadania - poprawy percepcji/ładności obrazu, a nie tylko metryk
%w kontekscie rozpoznawania, metryki?
%luka w obszarze wpływu na recognition w sporcie

\section{Methods}

\subsection{Data}

Our study utilizes a comprehensive dataset sourced from SoccerNet~\cite{soccernet2}, comprising 12 complete football games, all captured from the main camera perspective. This dataset is uniquely structured for the detailed analysis of player dynamics and interactions on the field. It consists of 100 video clips, each lasting 30 seconds, recorded from the main camera at a high-resolution quality of 1080p (1920×1080). These clips come from a diverse range of games and seasons, enabling a comprehensive evaluation and comparison of object detection methodologies across various match settings. This tracking dataset was divided by the authors into a training dataset consisting of 42.750 frames and a test dataset comprising 36.750 frames. In our analyses, we also adhered to this division, utilizing the training set for training purposes and conducting all tests on the test set as defined by the authors.

A critical aspect of this dataset is the categorization of on-field elements into distinct classes, which are crucial for object detection tasks. These classes include 'player team left', 'player team right', 'goalkeeper team left', 'goalkeeper team right', 'main referee', 'side referee', 'staff', and 'ball'. This classification enables a more nuanced and targeted approach to object detection, facilitating a detailed analysis of each entity's role and movement during the game. However, we have simplified it in our experiments by categorizing them into two groups: 'ball' and 'person'.

For the training of our super-resolution model, in addition to trainset of SoccerNet, we 
%incorporated two distinct datasets to evaluate the impact of dataset type on the accuracy of image reconstruction. The first dataset, DIV2K~\cite{li2022ntire}, is notable for its high-quality, diverse 2K resolution images. It comprises 1,000 images, with 800 designated for training, 100 for validation, and 100 for testing. This dataset is recognized for its utility in super-resolution research, offering a variety of image scenes and types of degradation, which includes standard bicubic downsampling and more complex degradations. The second dataset we 
used UHDSR8K~\cite{UHDSR8K}, which contains 2,100 images in 4K resolution. This dataset's high resolution offers a different range of challenges and opportunities for super-resolution model training. The UHDSR8K dataset is part of a study that aimed to benchmark single image super-resolution (SISR) methods on Ultra-High-Definition (UHD) images, including both 4K and 8K resolutions. It was used to evaluate the performance of SISR methods under various settings, contributing to the development of baseline models for super-resolution.  In our experiments, we specifically utilized the 4K images from this dataset.

%\subsection{Models}
\subsection{Super-resolution}
% \paragraph{super resolution}

\begin{table*}[htbp]
\caption{Impact of the shape of the input image on detection performance.}
\begin{center}
\begin{tabular}{|c|c|c|c|c|c|c|}
\hline
\textbf{Detector} & \textbf{Image Shape} & \multicolumn{2}{c|}{\textbf{mAP}} & \multicolumn{3}{c|}{\textbf{mean IoU}} \\
\cline{3-7} 
& & \textbf{@IoU=0.50:0.95} & \textbf{@IoU=0.50} & \textbf{$\tau_{IoU}=0.5$} & \textbf{$\tau_{IoU}=0.7$} & \textbf{$\tau_{IoU}=0.9$} \\
% \textbf{Detector} & \textbf{Image Shape} & \textbf{mAP@IoU=0.50:0.95} & \textbf{mAP@IoU=0.50}  & \textbf{mean IoU [$\tau_{IoU}=0.5$]} & \textbf{mean IoU [$\tau_{IoU}=0.7$]} & \textbf{mean IoU [$\tau_{IoU}=0.9$]} %\textbf{mAP-s} & \textbf{mAP-m} & \textbf{mAP-l}
\hline
 \multirow{5}{*}{Faster R-CNN} & QVGA (320×240) & 24.3 & 46.7 & 78.7 & 83.2 & 92.7  \\
 & VGA (640×480) & 27.6 &52.3 & 79.6&83.9 & 92.8 \\
 & DVD (720×576) & 27.5 & 51.8 & 79.6 & 83.9 & 92.8 \\
 & 720p (1280×720) &29.5&56.3 & 79.9&84.1 & 92.9  \\
 & Full HD (1920×1080) & 29.5 & 56.0 & 79.9 & 84.1 & 92.9 \\
 % \hline
 % \multirow{5}{*}{DETR}& QVGA (320×240)& & & &  \\
 % & VGA (640×480) & & & & \\
 % & DVD (720×576) & & & & \\
 % & 720p (1280×720) & & & & \\
 % & Full HD (1920×1080) & & & & \\
\hline
\end{tabular}
\label{tab:model-experiments-input-shape}
\end{center}
\end{table*}

Choosing the super-resolution network architecture for image reconstruction affects its quality, speed, and efficiency. The architecture selection in this study was based on the NTIRE 2022 Challenge on Efficient super-resolution results~\cite{li2022ntire}. The challenge required participating research groups to design a network for reconstructing a single image into higher resolution, aiming to improve network efficiency as measured by various metrics like runtime, parameter count, floating-point operations, and memory usage. The goal was to enhance efficiency while at least maintaining a PSNR (Peak Signal-to-Noise Ratio)~\cite{psnr} value 
29 on the DIV2K validation set. PSNR is a quality measure of image reconstruction compared to its original version, indicating lower distortion and information loss with higher values. The winning team, ByteESR, presented a network architecture named Residual Local Feature Network (RLFN)~\cite{rlfn}, which was chosen for this study due to its effective high-resolution image reconstruction and ability to preserve essential features and details with a relatively small size and quick learning process. The RLFN architecture stands out with its structure combining convolutional layers with attention mechanisms, allowing effective extraction of local image features and maintaining global pixel relations. This capability is crucial for precisely reflecting image structures and details in high-resolution image reconstruction.
Additionally, the RLFN network architecture is notable not just for its advanced capabilities but also for its relative simplicity, making it suitable for training and experimentation in academic settings. %Its efficiency on available hardware makes it especially valuable for academic research projects.

%We also conducted experiments with the Stable Diffusion x4 model~\cite{stable-diffusion-x4}, which is a specialized version of the Stable Diffusion series, designed for upscaling images. It is trained on a subset of the LAION dataset, specifically on images larger than 2048x2048 pixels, making it adept at enhancing lower-resolution images to higher resolutions. The model utilizes latent diffusion techniques for upscaling and is capable of processing images with a variety of content, including text prompts. 

\subsection{
%\paragraph{
Object detection} 
We used %two models for object detection: Mask R-CNN with a ResNet50 backbone~\cite{{maskrcnn}} and 
Faster-RCNN~\cite{fasterrcnn} with ResNet50 backbone for object detection. This choice was heavily influenced by the availability of pre-trained models in the PyTorch framework as well as the model's widespread adoption, which eased the process of adapting it to the project's specific needs. Additionally, the robustness and efficiency of Faster-RCNN in handling diverse datasets further justified its selection, ensuring reliable performance across varying object detection scenarios. %Both the Mask R-CNN and Faster R-CNN models underwent two types of evaluation of pretrained models on ... dataset: (1) without finetuning, and (2) with fine-tuning on SoccerNet.  This fine-tuning %was/
%could be a critical step in adjusting the pre-trained models to align more closely with the project's specific requirements, thereby enhancing their performance. 

\subsection{
%\paragraph{
Experimental Setup}

In the input dataset SoccerNet, clips are represented as images from successive frames, and such images were used in our experiments. These images have dimensions of 1920×1080, and upon degrading their quality to low resolution, we altered their dimensions by a factor of 2 (960x540), 3 (640x360), 4 (480x270), and 6 (320x180) to be inputed to super-resolution network. For UHDSR8K shape of the original images is (3840×1920), so we reduced the dimensions by a factor of 2 (1920x960), 3 (1280x640), 4 (960x480), and 6 (640x320).

In our final experiments, we compare the quality of object detection model (Faster R-CNN) on low-resolution images and on images whose quality has been enhanced using super-resolution network (RLFN) trained on SoccerNet and UHDSR8K. The visualization of the experiments is presented in Fig.~\ref{fig:experiment-diagram}. For comparing the quality of the detectors, the IoU and mAP metrics were used.

Experiments were conducted using two NVIDIA A100-40GB GPUs. Training of super-resolution models and object detection was performed with a batch size of 32. The super-resolution model was trained for 15.000 epochs with a learning rate changing through training from 1e-5 to 0.01.  %In the case of object detection, we trained models over 5 epochs on the training part of SoccerNet with a learning rate of 0.001. In both cases, 
We used the Adam optimizer.

\section{Results}

\begin{figure*}[!h]
    \centering
    \includegraphics[width=0.85\textwidth]{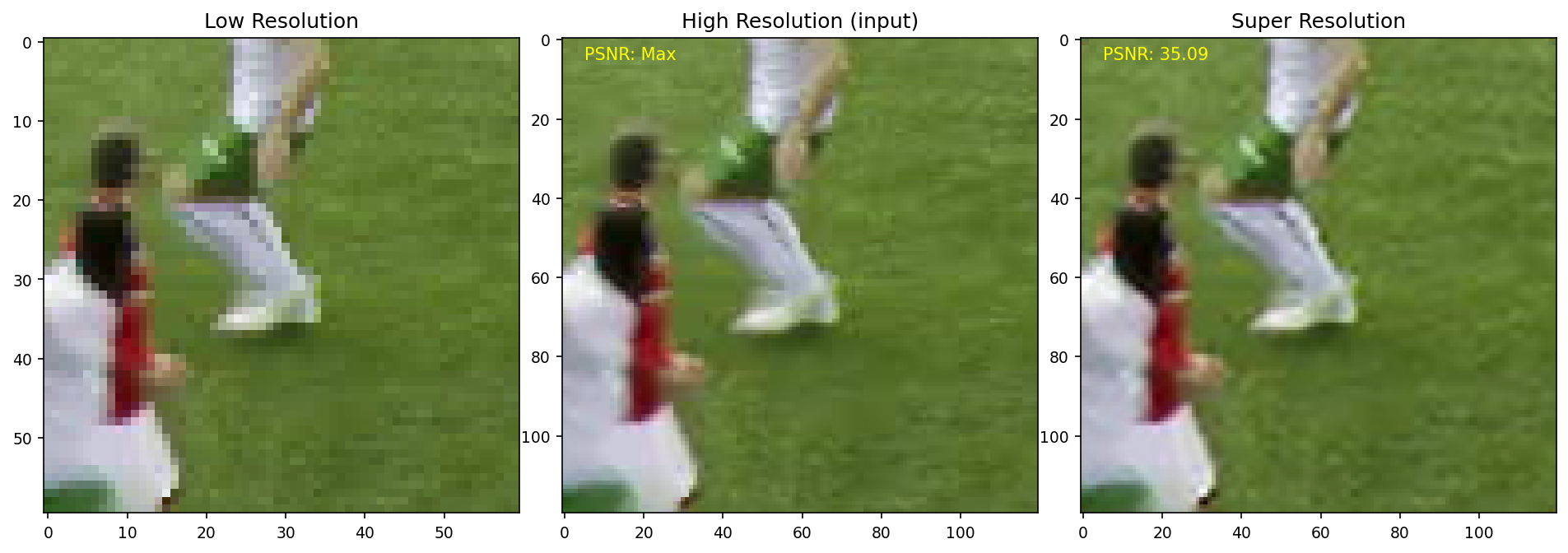}
    \hfill
    \includegraphics[width=0.85\textwidth]{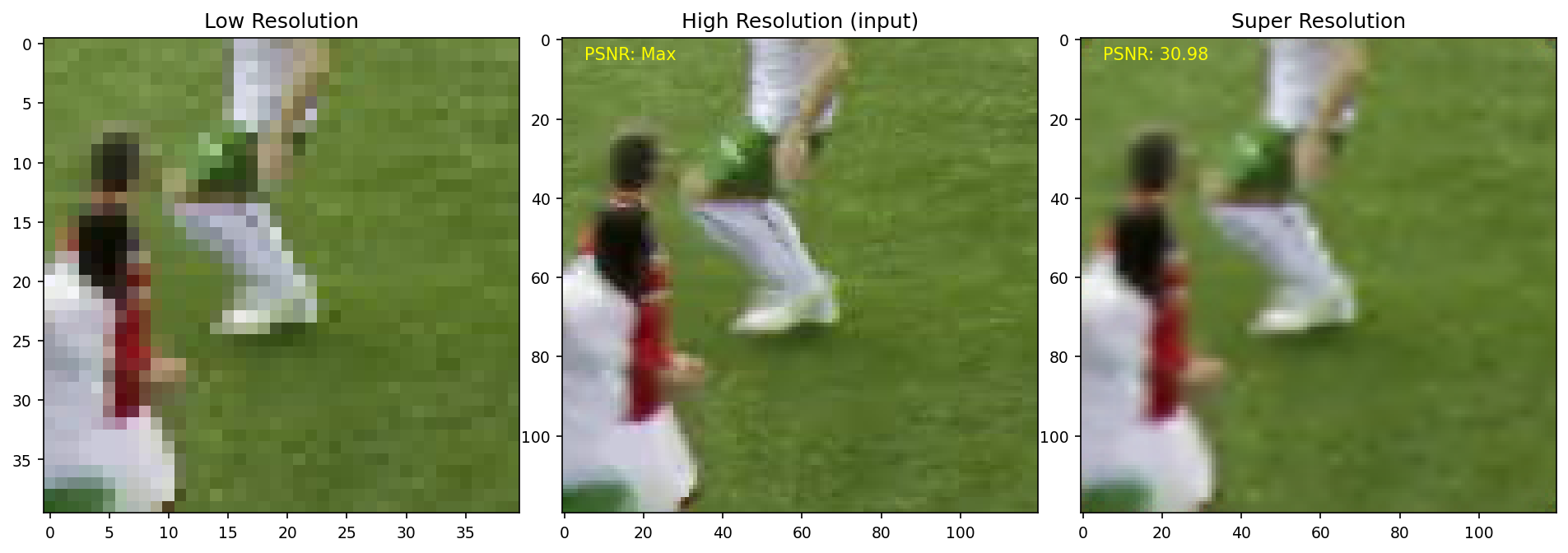}
    \hfill
    \includegraphics[width=0.85\textwidth]{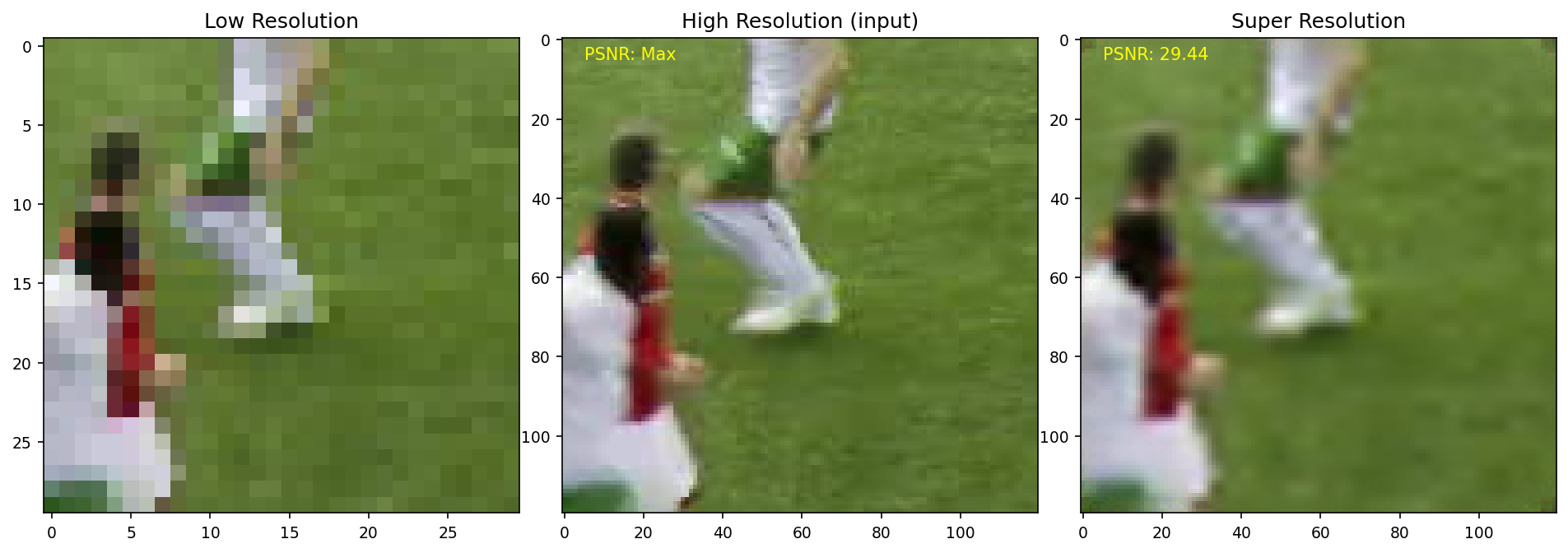}
    \hfill
    \includegraphics[width=0.85\textwidth]{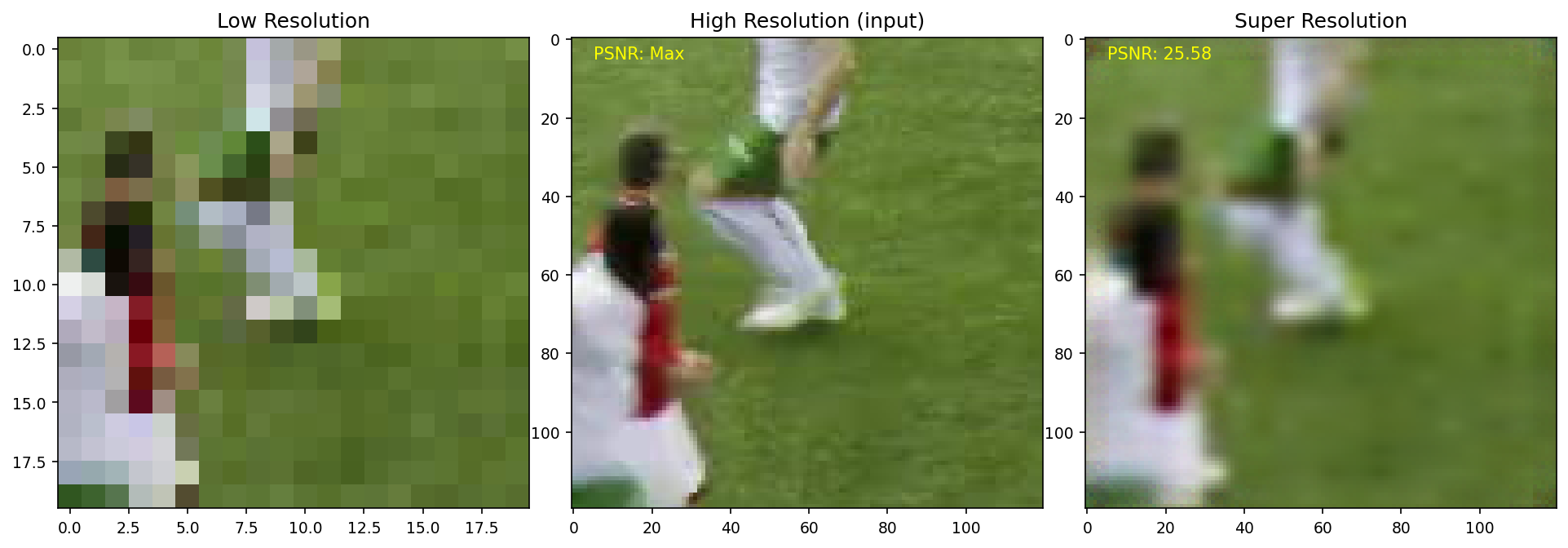}
    \caption{Sequential visualizations of super-resolution enhancements showcasing progressive upscaling factors. Rows of images illustrate the enhancements achieved through super-resolution RLFN techniques for scales of x2, x3, x4, and x6, respectively.}
    \label{fig:sr_examples}
\end{figure*}

In this section, we present the findings from our evaluation of how super-resolution methods affect the accuracy and reliability of object detection in football images. 

% Wyniki bez dotrenowania object detector

Table~\ref{tab:model-experiments-input-shape} presents the results of Faster R-CNN model tested on % which was fine-tuned on the training part of 
SoccerNet. % over five epochs.  
It is evident that an increase in the input size correlates with an enhancement in the metric mAP (mean Average Precision). Modifying the input size 6 times (from  320×240 to 1920×1080) resulted in an improvement of over 21\%  in mAP at an IoU (Intersection over Union) range of 0.50:0.95. Similar results, approximately a 20\% increase, were observed for mAP@IoU=0.50. Additionally, the average IoU was calculated across various IoU thresholds ($\tau_{IoU}$), yet the outcomes were relatively consistent across these measurements.

\begin{table}[!htbp]
\caption{Quality of super-resolution models on SoccerNet~\cite{soccernet2}. The notation 'x[N]' indicates that the image's dimensions were downscaled by a factor of N prior to undergoing enhancement through the super-resolution network. $\uparrow$ - the higher the better, $\downarrow$ - the lower the better}
\begin{center}
\begin{tabular}{|c|c|c|c|c|}
\hline
\textbf{Train Dataset} & \textbf{SR}  & \textbf{PSNR $\uparrow$}  & \textbf{MSE $\downarrow$} %& \textbf{SSIM} 
\\
\hline
 UHDSR8K & RLFN x2 & 33.15 & 36.47  \\
 UHDSR8K & RLFN x3 & 29.73 & 79.43 \\
 UHDSR8K & RLFN x4  & 27.83 & 119.82  \\
 UHDSR8K & RLFN x6  & 25.48 & 204.47  \\
 SoccerNet & RLFN x2 & 34.63 & 26.44  \\
 SoccerNet& RLFN x3 & 30.54 & 66.76  \\
 SoccerNet& RLFN x4  & 29.02 & 92.82  \\
 SoccerNet& RLFN x6  & 26.45 & 166.41  \\
 %LAION & Stable Diffusion x4 &  7.45& 12211.32  \\
 \hline
\end{tabular}
\label{tab:model-sr-quality}
\end{center}
\end{table}

% a PSNR above 30 dB is seen as indicative of good quality
% Limitations: PSNR assumes that the noise in the image is Gaussian and does not take into account other types of distortions specific to super-resolution processes, such as blurring or the introduction of unrealistic textures.

After training super-resolution models we evaluated their performance on SoccerNet using PSNR and MSE metrics. Table~\ref{tab:model-sr-quality} presents the quality of analyzed super-resolution methods. The degree of size reduction applied to the images before being inputted into the super-resolution model is denoted as x{N}. The metrics were calculated by comparing these images to their original versions. It is evident that greater size reduction correlates with increased difficulty in image reconstruction. Additionally, when the RLFN was trained on the SoccerNet training dataset, it showed a higher PSNR value on the test set of SoccerNet, which suggests that it is beneficial to train the network on images similar to the target dataset in order to achieve the best results. A PSNR value exceeding 30 is generally regarded as a sign of good quality. Therefore, in this instance, only the RLFN x2 and x3 models trained on SoccerNet, along with the RLFN x2 trained on UHDSR8K, can be considered as methods of good quality.

\begin{table*}[htbp]
\caption{Detection performance of Faster R-CNN model for various super-resolution scales and training datasets.}
\begin{center}
\begin{tabular}{|c|c|c|c|c|c|c|c|c|} %c|c|c|}
\hline
 \textbf{} &\textbf{} &\textbf{} &  & & \multicolumn{2}{c|}{\textbf{mAP}} & \multicolumn{2}{c|}{\textbf{mean IoU}} \\
\cline{4-9} 
%\textbf{SR Train} & 
\textbf{Input Shape} & \textbf{Train Dataset}& \textbf{Super-Resolution} & 
%\textbf{Input Shape} & 
\textbf{Output Shape}& \textbf{Scale} & \textbf{@IoU=0.50:0.95} & \textbf{@IoU=0.50} & \textbf{$\tau_{IoU}=0.5$} & \textbf{$\tau_{IoU}=0.9$}\\ %& \textbf{mAP-s} & \textbf{mAP-m} & \textbf{mAP-l}\\
\hline
 % \multirow{12}{*}{SoccerNet} & - & 720p (1280×720)& - & - & &  & & \\
 % %& RLFN & 720p (1280×720) & Full HD (1920×1080) & x1.5 & & & & &\\  % blad z wartoscia x1.5
 % %& Stable Diffusion & 720p (1280×720) & Full HD (1920×1080) & x2 & & & & &\\
 % & RLFN & 720p (1280×720) & 4K (3840×2160) & x3 & &  & &\\
 % %& RT4KSR & 720p (1280×720) & 4K (3840×2160) & x3 & & & & &\\
 % & RLFN & 720p (1280×720) & 8K (7680×4320) & x6 & &  & &\\
 %& Stable Diffusion & 720p (1280×720) & 8K (7680×4320) & x4 & & & & &\\

 % \cline{2-7}
 % & - & Full HD (1920×1080) & -  & - & &  \\
 % & RLFN & Full HD (1920×1080) & 4K (3840×2160)  & x2 & &  \\
 % %& RT4KSR & Full HD (1920×1080) & 4K (3840×2160)  & x2 & & & & &\\
 % %& Stable Diffusion & Full HD (1920×1080) & 4K (3840×2160)  & x2 & & & & &\\
 % & RLFN & Full HD (1920×1080) & 8K (7680×4320)  & x4 & &  \\
 % & Stable Diffusion & Full HD (1920×1080) & 8K (7680×4320)  & x4 & &  \\

  %\cline{2-9}
  \multirow{9}{*}{(320, 240)} & - &-  & -  & - & 24.3 & \textbf{46.7} & 78.7 & 92.7 \\
  \cline{2-9}
   &\multirow{4}{*}{SoccerNet}&RLFN & (640, 480)  & x2 & 26.0 & 39.8 & 81.1&91.9 \\
  & &RLFN &  (960, 720)  & x3 & 26.9 & 40.8 & 81.6&  92.6 \\
   &&RLFN &   (1280, 960)  & x4 & \textbf{27.3}& 41.4& 81.8&92.7 \\
  &&RLFN &   (1920, 1440) & x6 & 26.6&40.8 & 81.1 & \textbf{93.1} \\
    \cline{2-9}
&\multirow{4}{*}{UHDSR8K} &RLFN & (640, 480)  & x2 & 24.7 & 40.2  & 80.2 & 92.2 \\
  &&RLFN &    (960, 720)  & x3 & 24.9  & 39.7  & 80.4 & 91.7 \\
   &&RLFN &  (1280, 960)  & x4 & 25.3  & 40.3 &  80.8& 91.8\\
  &&RLFN &   (1920, 1440) & x6 &  27.2 & 41.3 & \textbf{82.1} & 92.5\\

  \cline{1-9}
  \cline{1-9}
  \multirow{9}{*}{(640, 480)} &- &- & -  & - &  27.6 & \textbf{52.3} & 79.6 & 92.8 \\
  \cline{2-9}
  &\multirow{4}{*}{SoccerNet} &RLFN &  (1280, 960) & x2 & 27.8 & 40.1 & 83.7& 92.9 \\
  &&RLFN &    (1920, 1440)  & x3 & 28.1 & 40.3 & 83.9 & 92.7 \\
  &&RLFN & (2560, 1920)  & x4 & \textbf{28.5} & 40.5 & 84.0& \textbf{93.0}\\
 &&RLFN &   (3840, 2880)  & x6 & 28.3 &40.4 & 83.8& 92.8\\
\cline{2-9}
 &\multirow{4}{*}{UHDSR8K} &RLFN ) &  (1280, 960) & x2 & 28.3 & 40.9 & 83.7& 92.9\\
  &&RLFN &   (1920, 1440)  & x3 & 28.0  & 40.1  & 84.1& 92.8\\
  &&RLFN &   (2560, 1920)  & x4 & \textbf{28.5}  & 40.3  & 84.2& \textbf{93.0}\\
 &&RLFN &  (3840, 2880)  & x6 & 28.4  &40.0  & \textbf{84.3}& 92.8\\

 \cline{1-9}
  \cline{1-9}
  \multirow{9}{*}{(1280, 720)} &- &- & -  & - & 29.5  & \textbf{56.3} & 79.9 & 92.9 \\
  \cline{2-9}
  &\multirow{4}{*}{SoccerNet}&RLFN  & (2560, 1440)  & x2 & 30.1  & 42.0  & 84.5& 92.9 \\
  &&RLFN &    (3840, 2160)  & x3 &  30.1&  41.8  &  84.2 & 92.9\\
  &&RLFN &   (5120, 2880)  & x4 & 30.3 & 42.4 & 84.2&  92.9\\
 &&RLFN &   (7680, 4320)  & x6 & 29.8 & 41.8& 84.2& 93.0\\
\cline{2-9}
  &\multirow{4}{*}{UHDSR8K}&RLFN  & (2560, 1440)  & x2 & 30.4 & 42.8 & 84.4& 92.8 \\
  &&RLFN &   (3840, 2160)  & x3 & 30.3  & 42.8 & 84.3& 93.0\\
  &&RLFN &  (5120, 2880)  & x4 & \textbf{30.6} & 42.6 & \textbf{84.8} & 92.9\\
 &&RLFN &  (7680, 4320)  & x6 & 29.6  & 41.8  & 84.3& \textbf{93.1}\\
\hline
\end{tabular}
\label{tab:model-results-without-training}
\end{center}
\end{table*}

Table~\ref{tab:model-results-without-training} presents the findings from our investigation into the effects of super-resolution techniques on object detection. We employed an object detection model in its original form, without any additional training on the dataset under study. Our findings reveal that applying super-resolution to 320x240 images leads to an approximate 12\% improvement in the mAP @IoU=0.50:0.95 metric, while the mAP@IoU=0.50 metric shows a decline. A similar trend is observed with 640x480 images, where the mAP @IoU=0.50:0.95 metric increased from 27.6 to 28.5 and 1280x720, where we observe increase from 29.5 to 30.6. We also evaluated the average IoU values at various thresholds ($\tau_{IoU}$), discovering that super-resolution enhances the average IoU at both evaluated thresholds. 
This shows that the application of super-resolution also helps in more accurately determining bounding boxes. 

It is not invariably true that increasing the size of images and the upscale factor leads to enhanced detection outcomes. Notably, a sixfold increase in image dimension does not manifest a discernible enhancement in performance relative to a fourfold increase. This phenomenon may be caused by the potential for the super-resolution network to introduce distortions into the image, thereby complicating the task of object detection. Nonetheless, it is important to notice that the outcomes obtained with the application of super-resolution techniques still surpass those achieved in the absence of any image quality enhancement methods.

Notably, although Table~\ref{tab:model-sr-quality} suggests that super-resolution models trained on the SoccerNet training dataset have superior quality, the relation to their detection efficacy remains ambiguous. The performance of the detector remains comparable after being processed by super-resolution models trained on both datasets. Furthermore, when enlarging an input image of 320x240 dimensions by six times, it is evident that the super-resolution model developed on the UHDSR8K dataset effected a superior image transformation, achieving a mean Average Precision (mAP) of 27.2, as opposed to 26.6 achieved by the model trained on SoccerNet.

Interestingly, even for images of relatively high quality at 1280x720, we observe an improvement in quality. However, it is noticeable that the most significant increase in detection quality occurs with the smallest image size (320x240), where the increase in mAP@IoU=0.50:0.95 is 12.3\%. For medium-sized images, there is a 3.3\% increase, and for the largest image size, there is a 3.7\% increase.

% \textbf{mAP-s} & \textbf{mAP-m} & \textbf{mAP-l} nie maja sensu, bo w innych wielkosciach obrazkow cos innego rozumiemy przez duze/male obiekty

% 720p (1280×720)
% Full HD 1080p (1920×1080)
% 4K UHD (3840×2160)
% 8K UHD (7680×4320)

% 1080p -> 4K x2
% 720p -> 4K x3
% 720p -> 2880 x4 

% Wyniki z dotrenowaniem object detector

% OBRAZKI

Fig.~\ref{fig:sr_examples} presents the results of the RLFN model applied to a sample from SoccerNet test set. The middle column illustrates the original input image, which was resized by factors of 2, 3, 4, and 6, followed by processing through super-resolution network. With a small - double reduction, the network performs well in reconstructing the original image. A PSNR value over 35 indicates high quality of reconstruction. However, the smaller the dimensions of the low-resolution image, the more challenging it becomes to replicate the input. In the last case - a sixfold reduction - the PSNR drops to 25.58. This example demonstrates that the network is capable of fairly accurately reproducing objects despite having highly blurred pixels as input.

\section{Discussions and Conclusions}

The findings of our study underscore the pivotal role advanced super-resolution (SR) methods play in improving the performance of object detection models in the football domain. By employing SR technique, we have demonstrated an enhancement in the detection of smaller objects, such as balls and distant players, which are often challenging to identify due to low resolution and image degradation.

Integrating SR methods with conventional object detection algorithms, such as Faster R-CNN, has yielded significant improvements in detection accuracy (12\% of mAP increase for low-quality images). This synergy not only elevates the quality of input images through upscaling but also preserves critical features necessary for accurate object identification. %The resultant high-resolution images provide a richer set of data for the detection models, thereby reducing the likelihood of false negatives and improving the overall precision.
Furthermore, our study highlights the importance of selecting a suitable dataset for training super-resolution (SR) models. Our empirical analysis revealed that training the SR algorithm with football-specific data slightly enhanced the quality of super-resolution on SoccerNet, as compared to SR models that were trained on disparate datasets.

% Furthermore, our study highlights the importance of selecting an appropriate SR technique tailored to the specific requirements of football imagery. Factors such as the level of detail needed, the computational efficiency of the SR algorithm, and the compatibility with existing object detection frameworks are crucial in optimizing performance.

The practical implications of these enhancements are far-reaching. For instance, in automated video analysis for coaching and tactical evaluation, higher detection accuracy enables more detailed and nuanced analysis of players' movements and actions. Similarly, in the context of automated officiating and player tracking, improved object detection can contribute to more reliable and fair decision-making processes.  Furthermore, these improvements can lead to cost reductions as there is less need to invest in high-quality cameras; the super-resolution technology compensates by enhancing the quality of the images captured by more standard equipment.

Looking ahead, our research opens several avenues for future exploration. One promising direction is the investigation of real-time SR and object detection algorithms that can operate efficiently in a live broadcast environment. Another area of interest is the exploration of domain-specific SR techniques that are optimized for varying weather conditions and lighting environments typical of football matches.

In our future work, we also plan to explore various super-resolution techniques and object detection models. We aim to evaluate the effectiveness of cutting-edge methods in enhancing image quality and accurately identifying objects. Our goal is to find optimal solutions for diverse applications. What is more, we would like to analyse how fine-tuning object detectors and using sliding windows~\cite{sahi} affects detection quality.

In conclusion, the integration of advanced super-resolution methods into object detection frameworks presents a significant advancement in the analysis of football imagery. By addressing the challenges posed by low-resolution images, this approach enhances the accuracy of object detection, thereby offering valuable insights for both tactical analysis and the development of automated officiating systems.

% \subsection{Figures and Tables}
% \paragraph{Positioning Figures and Tables} 
% Fig.~\ref{fig}

% \begin{table*}[htbp]
% \caption{Table Type Styles}
% \begin{center}
% \begin{tabular}{|c|c|c|c|}
% \hline
% \textbf{Table}&\multicolumn{3}{|c|}{\textbf{Table Column Head}} \\
% \cline{2-4} 
% \textbf{Head} & \textbf{\textit{Table column subhead}}& \textbf{\textit{Subhead}}& \textbf{\textit{Subhead}} \\
% \hline
% copy& More table copy$^{\mathrm{a}}$& &  \\
% \hline
% \multicolumn{4}{l}{$^{\mathrm{a}}$Sample of a Table footnote.}
% \end{tabular}
% \label{tab1}
% \end{center}
% \end{table*}

% \section*{Acknowledgment}

% \section*{References}

\vspace{12pt}
\color{red}

\end{document}